\documentclass[sigconf]{acmart}
\AtBeginDocument{%
  }

\usepackage{algorithm}
\usepackage{algorithmic}
\usepackage{booktabs}       
\usepackage{amsfonts}       
\usepackage{nicefrac}       
\usepackage{microtype}      
\usepackage{xcolor}         
\usepackage{amsmath}
\usepackage{amsthm}
\usepackage{multirow}
\usepackage{colortbl}
\definecolor{lightpastelpurple}{rgb}{0.69, 0.61, 0.85}
\colorlet{LightGreen}{lightpastelpurple!40}
\usepackage{tabularx}
\usepackage{tikz}
\usepackage{commath}
\usepackage{pifont}
\usepackage{caption}
\usepackage{subcaption}
\usepackage{float}

\setcopyright{acmlicensed}
\copyrightyear{2018}
\acmYear{2025}
\acmDOI{XXXXXXX.XXXXXXX}

\acmConference[LAVA '25] {Proceedings of the 2nd International Workshop on Large Vision - Language Model Learning and Applications}{October 27--31, 2025}{Dublin, Ireland.}
\acmBooktitle{Proceedings of the 2nd International Workshop on Large Vision - Language Model Learning and Applications (LAVA '25), October 27--31, 2025, Dublin, Ireland}
\acmISBN{979-8-4007-1840-3/2025/10}
\acmDOI{10.1145/XXXXXX.XXXXXX}
\begin{document}

\title{Accelerating Conditional Prompt Learning via Masked Image Modeling for Vision-Language Models}

\author{Phuoc-Nguyen Bui}
\authornote{Both authors contributed equally to this research.}
\affiliation{%
  \institution{Convergence Research Institute, Sungkyunkwan University}
  \city{Suwon}
  \country{South Korea}
}

\author{Khanh-Binh Nguyen}
\authornotemark[1]
\affiliation{%
  \institution{School of Information Technology, Deakin University}
  \city{Geelong}
  \country{Australia}
  }

\author{Hyunseung Choo}
\authornote{Corresponding author}
\affiliation{%
  \institution{Department of Electrical and Computer Engineering, Sungkyunkwan University}
  \city{Suwon}
  \country{South Korea}
  }


\begin{abstract}
    Vision-language models (VLMs) like CLIP excel in zero-shot learning but often require resource-intensive training to adapt to new tasks. Prompt learning techniques, such as CoOp and CoCoOp, offer efficient adaptation but tend to overfit to known classes, limiting generalization to unseen categories. We introduce ProMIM, a plug-and-play framework that enhances conditional prompt learning by integrating masked image modeling (MIM) into existing VLM pipelines. ProMIM leverages a simple yet effective masking strategy to generate robust, instance-conditioned prompts, seamlessly augmenting methods like CoOp and CoCoOp without altering their core architectures. By masking only visible image patches and using these representations to guide prompt generation, ProMIM improves feature robustness and mitigates overfitting, all while introducing negligible additional computational cost. Extensive experiments across zero-shot and few-shot classification tasks demonstrate that ProMIM consistently boosts generalization performance when plugged into existing approaches, providing a practical, lightweight solution for real-world vision-language applications.
\end{abstract}





\maketitle

\section{Introduction}
\label{sec:intro}

Vision-language models (VLMs), such as Contrastive Language–\\Image Pretraining \cite{radford2021learning} (CLIP), A Large-scale ImaGe and Noisy-text embedding  \cite{jia2021scaling} (ALIGN), and Flamingo \cite{alayrac2022flamingo}, represent a significant leap in aligning visual and textual representations within a shared embedding space. These models enable zero-shot transfer learning by associating images with descriptive language labels, which provides valuable flexibility for downstream applications. However, training such models often requires large amounts of data and computing resources, which are available primarily to large institutions. For example, CLIP leverages an extensive dataset of image-text pairs, which makes it difficult for smaller research teams to replicate its scale and effectiveness. Furthermore, while large datasets enable high performance, they can also introduce biases or irrelevant information, limiting the generalizability of these models to niche or domain-specific applications.

\begin{figure}[t]
\centering
\includegraphics[width=0.85\linewidth]{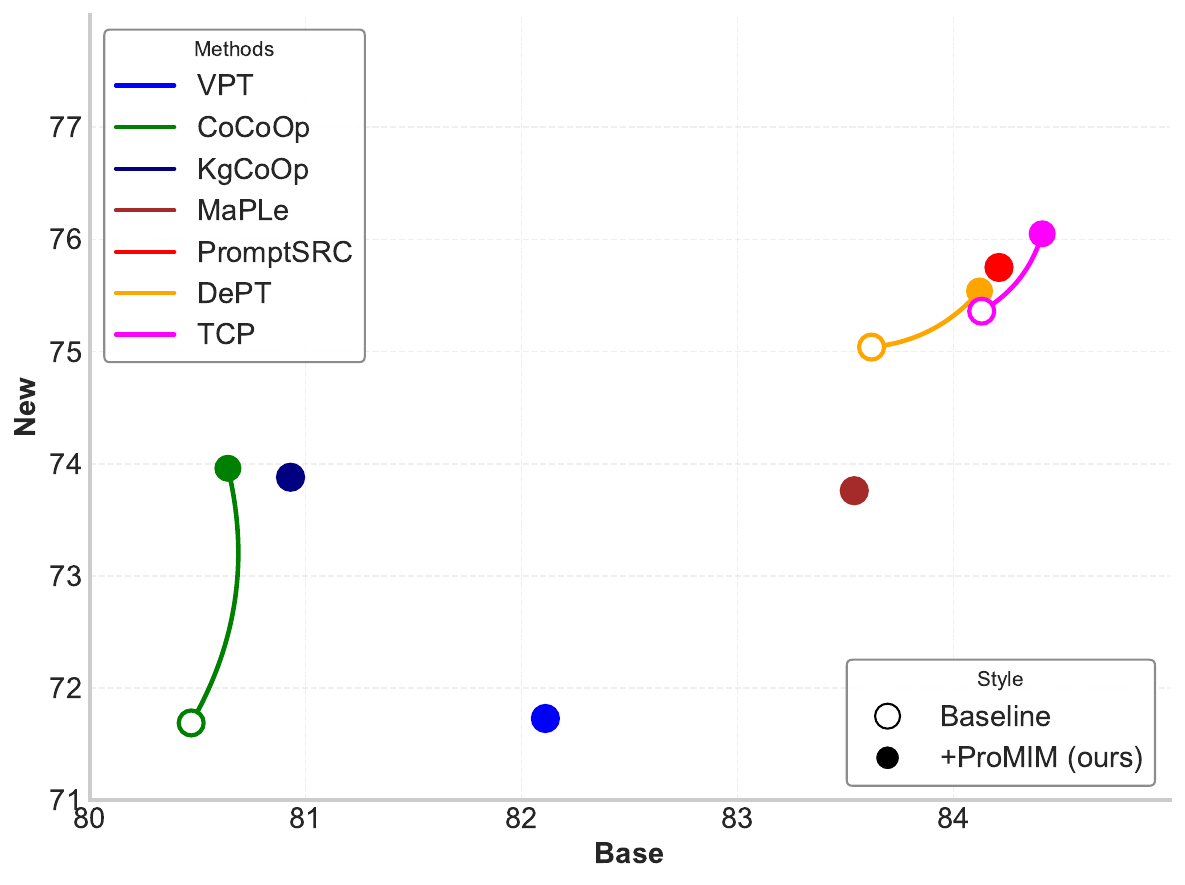} 
\caption{
Classification accuracies of six prompt tuning methods w/ or w/o our ProMIM on \textbf{Base} (or {seen}) and \textbf{New} ({or unseen}) tasks, averaged over 11 datasets in Table~\ref{table1}.
} 
\label{f1}
\end{figure}

\begin{figure}
    \centering
    \includegraphics[width=\linewidth]{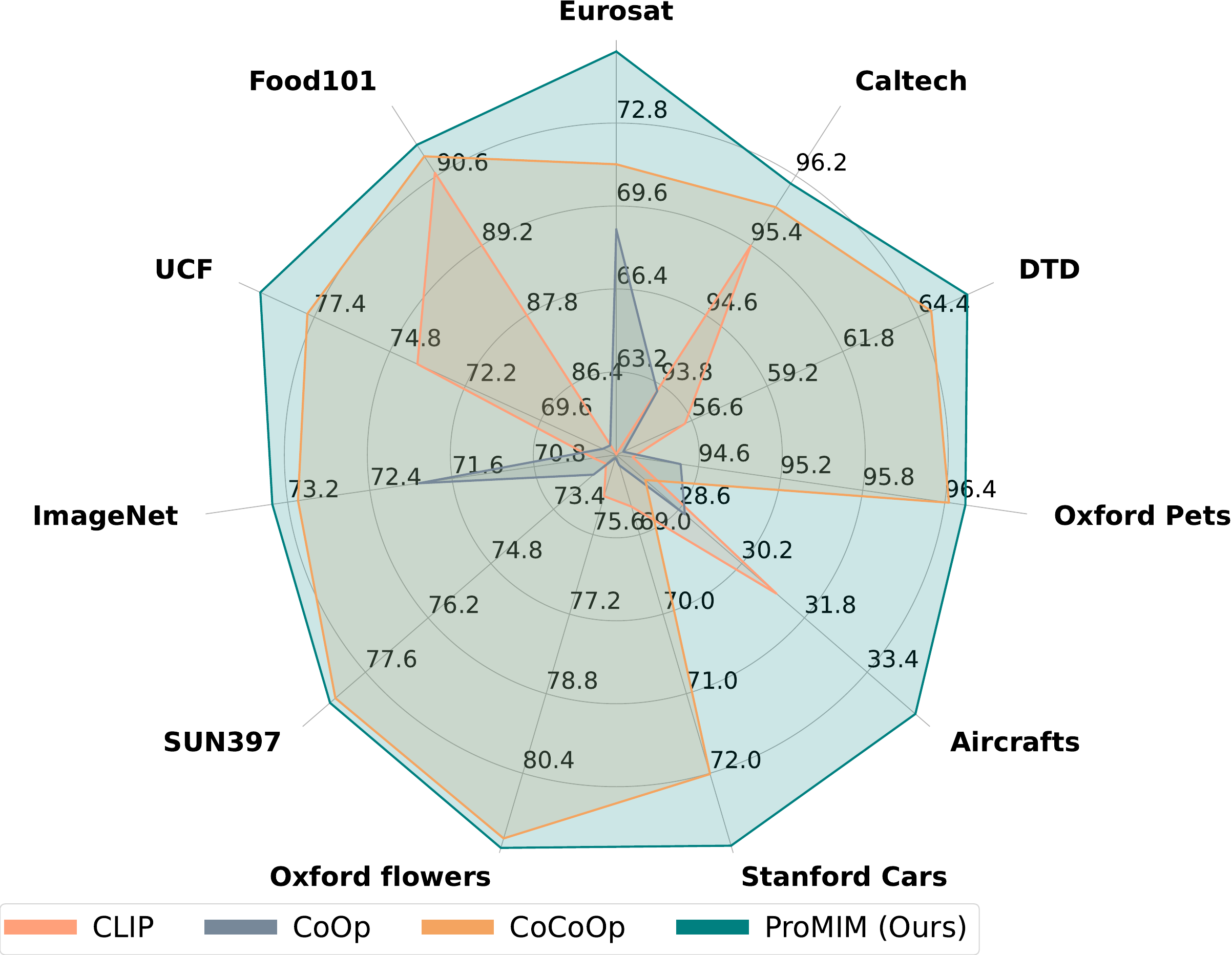}
    \caption{Performance comparison on base-to-novel generalization.}
    \label{fig:radar}
\end{figure}

Prompt learning has emerged as an efficient adaptation strategy for VLMs, inspired by prompt-tuning techniques in natural language processing (NLP) \cite{petroni2019language, cho2021unifying, tsimpoukelli2021multimodal, liu2023pre, bui2024visual}. Through prompt learning, these models can adapt more flexibly to new tasks by embedding additional context-specific information directly into the prompt rather than modifying the entire network. However, prompt-based methods face challenges, especially in maintaining performance on unseen classes. For instance, Context Optimization (CoOp) \cite{zhou2022learning} is prone to overfitting; while it performs well on seen classes, its accuracy drops significantly on unseen classes, often under-performing compared to CLIP’s zero-shot capabilities. This is because CoOp tends to specialize prompts for known classes, which limits model's generalization capability.

To address this, Zhou et al. \cite{zhou2022conditional} introduced Conditional Context Optimization (CoCoOp) method with instance-conditioned prompts, aiming to adapt prompts based on the specific features of each instance, enhancing CoOp’s flexibility. Although CoCoOp shows improved generalization on unseen classes, it still suffers from overfitting due to data leakage issues, where prompts trained on seen instances can inadvertently capture information that affects performance on unseen classes \cite{ma2022understanding}.
Recently, KgCoOp (Knowledge-guided Context Optimization) \cite{yao2023visual} was developed to tackle this issue by aligning learnable prompts more closely with hand-crafted prompts, thereby reducing context discrepancies, enhancing prompt stability across diverse tasks and reducing the overfitting problem. However, KgCoOp’s improvements are primarily confined to text features, without fully leveraging the vision encoder’s potential. This lack of integration with visual features limits its capacity to harness the vision-language model’s full representational power.

To address these limitations, we propose Masked Image Modeling-guided Conditional Prompt Learning (ProMIM), a plug-and-play enhancement for prompt learning in VLMs that integrates masked image modeling (MIM) to boost generalization without disrupting existing frameworks as shown in Figure~\ref{f1}. Designed to seamlessly augment methods like CoOp and CoCoOp, ProMIM acts as a modular add-on that enhances their performance while preserving compatibility with pretrained models. By randomly masking image patches and using these representations to guide prompt generation, ProMIM encourages the model to learn robust, domain-invariant features, reducing overfitting to specific visual details. This lightweight approach requires no architectural overhaul—ProMIM simply enhances the input processing stage, making it an easy-to-adopt solution for adapting VLMs. Unlike KgCoOp’s text-centric design, ProMIM leverages masked visual information, unlocking the vision encoder’s full representational power while integrating effortlessly with existing pipelines. As a result, ProMIM improves adaptability to unseen data and maintains strong performance on seen classes, offering a practical tool for real-world vision-language tasks. The contributions are as follows:

\begin{itemize}
    \item We present ProMIM, a plug-and-play framework that enhances conditional prompt learning by incorporating masked image modeling, providing a drop-in solution to improve generalization in VLMs.
    \item Through comprehensive experiments on different configurations, we show that ProMIM consistently elevates the performance of existing CoOp-based methods, achieving superior adaptability to unseen classes.
    \item We validate the robustness and flexibility of ProMIM across a wide range of datasets, highlighting its potential as a versatile, easy-to-adopt tool for vision-language applications requiring strong generalization.
\end{itemize}

\section{Related Work}
\label{sec:related_work}
\subsection{Visual-Language Models}
Recent research highlights that pairing images with associated text, rather than analyzing images alone, can lead to highly effective vision-language models (VLMs). A key example of this approach is CLIP \cite{radford2021learning}, which uses contrastive loss to train both a vision encoder and a text encoder on a dataset of 400 million image-text pairs, demonstrating impressive generalization to unseen classes. Vision-language models like CLIP harness the relationship between images and text, opening up new possibilities for learning generalized visual representations. VLMs have continued to evolve, improving through the use of more advanced text and visual encoders, such as Transformers \cite{dosovitskiy2020image}; contrastive learning techniques \cite{chen2020simple}; and increasingly large datasets \cite{jia2021scaling}. Since training VLMs typically requires extensive annotated datasets, unsupervised and weakly supervised methods \cite{wang2021simvlm} have been explored as alternatives to annotate-free training. Specifically, Masked Language Modeling (MLM) \cite{kim2021vilt, lu2019vilbert} enhances robustness in text and visual embeddings by randomly masking words in text, while Masked Auto-Encoders (MAE) employ self-supervised learning by masking random image patches, facilitating the development of adaptable, domain-agnostic representations that enhance the model’s generalization capacity across diverse tasks and domains.

\subsection{Prompt Tuning} 

To customize pretrained VLMs for downstream tasks, prompt tuning \cite{petroni2019language} integrates task-specific textual tokens to extract relevant knowledge for each task \cite{tsimpoukelli2021multimodal}. For instance, CLIP \cite{radford2021learning} uses a manually crafted template, such as ``a photo of a [CLS],'' to create textual embeddings for zero-shot inference. However, such hand-crafted prompts can lack expressiveness, as they may not fully capture task-specific nuances. To address this, Context Optimization (CoOp) \cite{zhou2022learning} replaces these static prompts with learnable, soft prompts based on labeled few-shot samples, enhancing adaptability to specific tasks. A limitation of CoOp, however, is that its learnable prompts remain fixed across all images within a task, thus overlooking individual image characteristics. Conditional Context Optimization (CoCoOp) \cite{zhou2022conditional} advances this by generating an image-conditioned prompt for each instance, which it combines with a textual-conditioned context to create task-specific prompts. This approach employs a lightweight neural network to generate an adaptive, learnable text prompt for each image.

Beyond text-only prompt tuning, approaches such as Multimodal Prompt Learning (MaPLe) \cite{khattak2023maple} and PromptSRC \cite{khattak2023self} incorporate both visual and textual prompt tuning, optimizing jointly across the vision and text encoders. Furthermore, Multi-task Vision-Language Prompt Tuning (MVIPT) \cite{shen2024multitask} integrates cross-task knowledge to enhance prompt learning for VLMs, while DenseCLIP \cite{rao2022denseclip} employs context-aware prompt adjustments tailored for dense prediction tasks. Another approach, CLIP-Adapter \cite{gao2024clip}, enhances adaptation by introducing adapters that fine-tune both visual and textual embeddings, thus enabling VLMs to better accommodate diverse downstream applications. Recently, CasPL \cite{wu2024cascade} introduces a plug-and-play, two-phase framework—boosting and adapting prompts—to improve generalization with minimal overhead. Unlike CasPL’s two-phase approach, our ProMIM offers a one-phase, plug-and-play method with masked image modeling, building on CoCoOp to enhance robustness and generalization efficiently. 


\begin{figure*}[tp]
    \centering
    \includegraphics[width=0.7\linewidth]{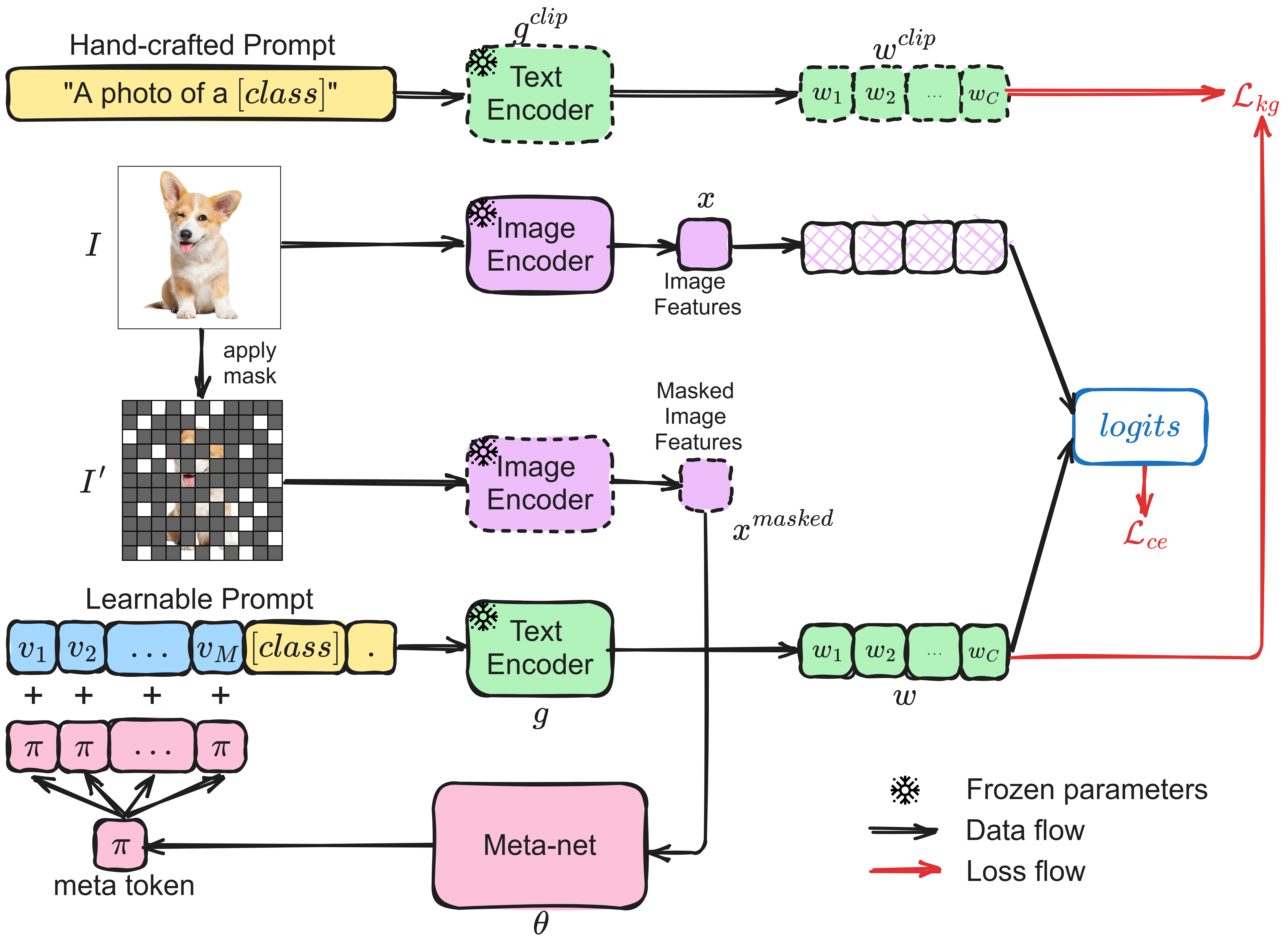}
    \caption{The framework of the proposed ProMIM.
    The masked input image embedding is used to generate the learnable meta token that integrates with the learnable context tokens. Thus, ProMIM mitigates the overfitting caused by data leakage.}
    \label{fig:pipeline}
\end{figure*}

\section{Methodology}
\subsection{Prompt learning for CLIP}
Prompt learning removes the dependency on manually crafted prompts, such as 'a photo of a [...]' to better align with downstream task requirements. The pioneering work, CoOp \cite{zhou2022learning}, defines a prompt as a sequence of $M$ continuously differentiable tokens, denoted by $[v_1][v_2] \dots [v_M]$. In the case of the CLIP-ViT architecture, each token $[v_i]$ is represented as a 512-dimensional vector. Consequently, the prompt for the $i$-th class can be formulated as $t_i(x) = {v_1(x), v_2(x), \dots, v_M(x), c_i}$, where $c_i$ is the class label. Let $x$ represent the feature embeddings from the image encoder, and $g(\cdot)$ denote the features from the text encoder. The class probabilities are then computed as follows:
\begin{equation}
    p(\hat{y} \mid x)=\frac{\exp \left(\operatorname{sim}\left(x, g\left(\mathbf{t}_y\right)\right) / \tau\right)}{\sum_{i=1}^C \exp \left(\operatorname{sim}\left(x, g\left(\mathbf{t}_i\right)\right) / \tau\right)}
\end{equation}
Here, $\text{sim}(\cdot, \cdot)$ denotes a similarity function within the feature space, where cosine similarity is frequently adopted, and $\tau$ represents the temperature parameter controlling the concentration of the distribution. Conditional Context Optimization (CoCoOp) \cite{zhou2022conditional} leverages image-conditioned prompts to improve generalization across unseen categories by dynamically adapting to input image features. Specifically, these image-conditioned prompts, $t_i(x) = {v_1(x), v_2(x), \dots, v_M(x), c_i}$, are computed by combining meta-tokens $\pi$ generated by a 'meta-network' $h_{\theta}$ with the token sequence $[v_i]$. The class probabilities are defined as:
\begin{equation}
    p(\hat{y} \mid x)=\frac{\exp \left(\operatorname{sim}\left(x, g\left(\mathbf{t}_y(x)\right)\right) / \tau\right)}{\sum_{i=1}^C \exp \left(\operatorname{sim}\left(x, g\left(\mathbf{t}_i(x)\right)\right) / \tau\right)}
\label{eq:eq2}
\end{equation}
Both CoOp and CoCoOp update token representations; however, CoCoOp further optimizes the meta-network parameters $h_{\theta}$ using cross-entropy loss derived from the downstream task objectives described as follows: 
\begin{equation}
    \mathcal{L}_{\text{ce}}(y,\hat{y})=-\sum_{i=1}^C y_i \text{log}\left(\hat{y}_i\right)
\end{equation}

\citet{yao2023visual} presents Knowledge-guided Context Optimization (KgCoOp) as a method to improve the generalizability of learnable prompts for classes not encountered during training.
This is achieved by reducing the difference between textual embeddings produced by the learned prompts and those created with handcrafted prompts. One of the primary limitations of current fine-tuning techniques is the substantial reduction in zero-shot generalization when newly introduced learnable parameters are specifically adapted for a downstream task, often resulting in overfitting to the downstream dataset. In the subsequent section, we present ProMIM, a novel framework designed to mitigate this issue by enhancing both downstream task performance and zero-shot prediction capabilities.

\subsection{ProMIM}
\paragraph{\textbf{Motivation}}
While CoCoOp demonstrates impressive performance, a notable limitation is its tendency to be overly confident with unfamiliar classes.
We conjecture that this challenge arises from utilizing image features created by the visual path, which are then transferred to the textual path to produce prompts. 
This technique increases the potential for data leakage because features from the visual side may leak to the textual side and lead the model to overfit to the training data, thereby diminishing its adaptability and ultimately compromising its generalizability.

A straightforward remedy is employing distinct encoders for various augmented input images to enhance diversity.
However, as demonstrated in \cite{li2024erroneousagreementsclipimage}, even with image flipping, CLIP maintains identical cosine similarity.
Consequently, utilizing different augmented input images with separate encoders is rendered ineffective.

\paragraph{\textbf{Masked Image Modeling as a regularization}}
ProMIM tackles the drawback of poor generalization due to overfitting on the downstream tasks by implementing a masked image modeling strategy to conventional prompt learning.
To prevent the data leakage issue from happening, we randomly mask the input image before inputting it to the image encoder.
ProMIM prevents the model from remembering the global information in training data, and the model is encouraged to learn more generalized features that are applicable to a wider range of images.


In addition, using masked image features to generate the context tokens, rather than relying on full image features, can enhance robustness. The masked features force the model to learn from incomplete information, making it more resilient to variations and occlusions in the image \cite{song2019occlusion, kong2023understanding}. By focusing on these masked features, the model is encouraged to capture the core, essential characteristics of objects and scenes, thereby improving its generalization ability to unseen classes. However, utilizing the same image encoder for both the full image and the masked image to generate context tokens is inefficient, as it requires processing the data twice—once for the full image and again for the masked image—before performing backward propagation. This redundant processing increases computational overhead and limits efficiency.

To address this issue, we utilize the powerful generalized CLIP as the individual encoder, as depicted in Figure~\ref{fig:pipeline}.
We use the pre-trained CLIP image encoder for image encoding.
Initially, the image $I$ is segmented into a grid consisting of non-overlapping patches.
A substantial number of these patches (e.g., 50\% or 75\%) are randomly masked to obtain the masked image $I'$, where the image encoder processes only the visible patches, following the approach in \cite{he2022masked}.
By applying a masking ratio of 50\% (or 75\%), the computational overhead for masked image encoding is reduced to half (or quarter), and this adjustment also allows the utilization of a batch size while maintaining a similar memory requirement for image encoding.

Let $h_{\theta}(\cdot)$ represents the meta-net parameterized by $\theta$, where each context token is now computed as $v_{m}(x^{\text{masked}}) = v_m + \pi$, with $\pi = h_{\theta}(x^{\text{masked}})$ and $m \in \{1, 2, \dots, M\}$. Consequently, the prompt for the $i$-th class becomes conditioned on the masked input, formulated as $t_i(x^{\text{masked}}) = \{v_1(x^{\text{masked}}), v_2(x^{\text{masked}}), \dots, v_M(x^{\text{masked}}), c_i\}$. During training, both the context vectors ${v_m}_{m=1}^{M}$ and the parameters $\theta$ of the meta-network are jointly updated. The prediction probability from Equation~\ref{eq:eq2} is then modified as follows:
\begin{equation}
    p(\hat{y} \mid x)=\frac{\exp \left(\operatorname{sim}\left(x, g\left(\mathbf{t}_y(x^{masked})\right)\right) / \tau\right)}{\sum_{i=1}^C \exp \left(\operatorname{sim}\left(x, g\left(\mathbf{t}_i(x^{masked})\right)\right) / \tau\right)}
\end{equation}

To strengthen regularization under the consistency constraint, we utilize the standard CLIP text encoder and apply regularization of KgCoOp.
We define the text embeddings produced by CLIP and ProMIM as $w^{clip}_i = g^{clip}\left(t^{clip}_i\right)$ and $w^{ProMIM}_i = g(t_i)$, respectively.
$t^{clip}_i=e(\text{“a photo of a [CLS]”})$ represents the vectorized textual tokens in CLIP of the i-th class template "a photo of a [CLS]", while $g^{clip}$ and $g$ denote the text encoders for the hand-crafted prompt and the soft prompt, respectively.
In line with KgCoOp, we aim to decrease the distance between $w^{ProMIM}_i$ and $w^{clip}_i$ to enhance the generability of the unseen classes.
\begin{equation}
    \mathcal{L}_{kg} = \frac{1}{N_c}\sum_{i=1}^{N_c}\norm{w^{ProMIM}_i - w^{clip}_i}_2^2
\end{equation}
where $\norm{\cdot}$ denotes the Euclidean distance, and $N_c$ represents the number of seen classes.
The final objective is:
\begin{equation}
    \mathcal{L} = \mathcal{L}_{ce} + \lambda \mathcal{L}_{kg}
\end{equation}
where $\lambda$ serves as a balancing factor for the influence of $\mathcal{L}_{kg}$ in the overall objective function.
 
\section{Experiments}
Our approach is comprehensively evaluated across three configurations:
1) generalization from base classes to novel classes within the same dataset,
2) cross-dataset transferability, and 3) domain generalization.
All models utilized in our experiments are implemented using the open-source CLIP framework \cite{radford2021learning}.

\begin{table*}[!htp]
\caption{Quantitative comparison with existing methods under the base-to-new generalization setting with ViT-B/16 as the backbone on 11 datasets. The context length $M$ is set to 4 for prompt-based approaches, with 16-shot samples drawn from the base classes. H: Harmonic mean.}
\label{table1}
\centering
\subfloat[\textbf{Average over 11 datasets.}]{
\resizebox{0.24\linewidth}{!}{
    \begin{tabular}{@{}lcc|c@{}}
    \toprule
    ViT-B/16 & Base & New & H \\
    \midrule
    CLIP & 69.34 & 74.22 & 71.70 \\
    \midrule
    CoOp & 82.69 & 63.22 & 71.66 \\
    KgCoOp & 80.73 & 73.60 & 77.00 \\
    ProGrad & 82.48 & 70.75 & 76.16 \\
    MaPLe & 82.28 & 75.14 & 78.55 \\
    PromptSRC & 84.26 & 76.10 & 79.97 \\
    \midrule
    CoCoOp & 80.47 & 71.69 & 75.83 \\
    \rowcolor{LightGreen}\textbf{+ProMIM} & \textbf{80.64} & \textbf{73.96} & \textbf{77.16} \\
    \midrule
    DePT & 83.62 & 75.04 & 79.10 \\
    \rowcolor{LightGreen}\textbf{+ProMIM} & \textbf{84.12} & \textbf{75.54} & \textbf{79.60} \\
    \midrule
    TCP & 84.13 & 75.36 & 79.51 \\
    \rowcolor{LightGreen}\textbf{+ProMIM} & \textbf{84.41} & \textbf{76.05} & \textbf{80.01} \\
    \bottomrule
    \end{tabular}
    }
}
\subfloat[ImageNet.]{
\resizebox{0.24\linewidth}{!}{
    \begin{tabular}{@{}lcc|c@{}}
    \toprule
    ViT-B/16 & Base & New & H \\
    \midrule
    CLIP & 72.43 & 68.14 & 70.22 \\
    \midrule
    CoOp & 76.47 & 67.88 & 71.92 \\
    KgCoOp & 75.83 & 69.96 & 72.78 \\
    ProGrad & 77.02 & 66.66 & 71.46 \\
    MaPLe & 76.66 & 70.74 & 73.47 \\
    PromptSRC & 77.60 & 70.73 & 74.01 \\
    \midrule
    CoCoOp & 75.98 & 70.43 & 73.10 \\
    \rowcolor{LightGreen}\textbf{+ProMIM} & \textbf{76.12} & \textbf{70.77} & \textbf{73.35} \\
    \midrule
    DePT & 77.03 & 70.13 & 73.42 \\
    \rowcolor{LightGreen}\textbf{+ProMIM} & \textbf{77.23} & \textbf{70.23} & \textbf{73.56} \\
    \midrule
    TCP & 77.27 & 69.87 & 73.38 \\
    \rowcolor{LightGreen}\textbf{+ProMIM} & \textbf{77.43} & \textbf{70.63} & \textbf{73.88} \\
    \bottomrule
    \end{tabular}
    }
}
\subfloat[Caltech.]{
\resizebox{0.24\linewidth}{!}{
    \begin{tabular}{@{}lcc|c@{}}
    \toprule
    ViT-B/16 & Base & New & H \\
    \midrule
    CLIP & 96.84 & 94.00 & 95.40 \\
    \midrule
    CoOp & 98.00 & 89.81 & 93.73 \\
    KgCoOp & 97.72 & 94.39 & 96.03 \\
    ProGrad & 98.02 & 93.89 & 95.91 \\
    MaPLe & 97.74 & 94.36 & 96.02 \\
    PromptSRC & 98.10 & 94.03 & 96.02 \\
    \midrule
    CoCoOp & 97.96 & 93.81 & 95.84 \\
    \rowcolor{LightGreen}\textbf{+ProMIM} & \textbf{98.03} & \textbf{94.27} & \textbf{96.11} \\
    \midrule
    DePT & 98.30 & 94.60 & 96.41 \\
    \rowcolor{LightGreen}\textbf{+ProMIM} & \textbf{98.32} & \textbf{94.67} & \textbf{96.46} \\
    \midrule
    TCP & 98.23 & 94.67 & 96.42 \\
    \rowcolor{LightGreen}\textbf{+ProMIM} & \textbf{98.34} & \textbf{94.89} & \textbf{96.58} \\
    \bottomrule
    \end{tabular}
    }
}
\subfloat[Pets.]{
\resizebox{0.24\linewidth}{!}{
    \begin{tabular}{@{}lcc|c@{}}
    \toprule
    ViT-B/16 & Base & New & H \\
    \midrule
    CLIP & 91.17 & 97.26 & 94.12 \\
    \midrule
    CoOp & 93.67 & 95.29 & 94.47 \\
    KgCoOp & 94.65 & 97.76 & 96.18 \\
    ProGrad & 95.07 & 97.63 & 96.33 \\
    MaPLe & 95.43 & 97.76 & 96.58 \\
    PromptSRC & 95.33 & 97.30 & 96.30 \\
    \midrule
    CoCoOp & 95.20 & 97.69 & 96.43 \\
    \rowcolor{LightGreen}\textbf{+ProMIM} & \textbf{95.37} & \textbf{97.76} & \textbf{96.55} \\
    \midrule
    DePT & 94.33 & 97.23 & 96.46 \\
    \rowcolor{LightGreen}\textbf{+ProMIM} & \textbf{95.22} & \textbf{97.73} & \textbf{95.78} \\
    \midrule
    TCP & 94.67 & 97.20 & 95.92 \\
    \rowcolor{LightGreen}\textbf{+ProMIM} & \textbf{94.87} & \textbf{97.53} & \textbf{96.18} \\
    \bottomrule
    \end{tabular}
    }
}
\quad%
\subfloat[Cars.]{
\resizebox{0.24\linewidth}{!}{
    \begin{tabular}{@{}lcc|c@{}}
    \toprule
    ViT-B/16 & Base & New & H \\
    \midrule
    CLIP & 63.37 & 74.89 & 68.65 \\
    \midrule
    CoOp & 78.12 & 60.40 & 68.13 \\
    KgCoOp & 71.76 & 75.04 & 73.36 \\
    ProGrad & 77.68 & 68.63 & 72.88 \\
    MaPLe & 72.94 & 74.00 & 73.47 \\
    PromptSRC & 78.27 & 74.97 & 76.58 \\
    \midrule
    CoCoOp & 70.49 & 73.59 & 72.01 \\
    \rowcolor{LightGreen}\textbf{+ProMIM} & \textbf{71.67} & \textbf{74.20} & \textbf{72.91} \\
    \midrule
    DePT & 79.13 & 75.47 & 77.26 \\
    \rowcolor{LightGreen}\textbf{+ProMIM} & \textbf{79.40} & \textbf{75.63} & \textbf{77.47} \\
    \midrule
    TCP & 80.80 & 74.13 & 77.32 \\
    \rowcolor{LightGreen}\textbf{+ProMIM} & \textbf{80.93} & \textbf{74.37} & \textbf{77.51} \\
    \bottomrule
    \end{tabular}
    }
}
\subfloat[Flowers.]{
\resizebox{0.24\linewidth}{!}{
    \begin{tabular}{@{}lcc|c@{}}
    \toprule
    ViT-B/16 & Base & New & H \\
    \midrule
    CLIP & 72.08 & 77.80 & 74.83 \\
    \midrule
    CoOp & 97.60 & 59.67 & 74.06 \\
    KgCoOp & 95.00 & 74.73 & 83.65 \\
    ProGrad & 95.54 & 71.87 & 82.03 \\
    MaPLe & 95.92 & 72.46 & 82.56 \\
    PromptSRC & 98.07 & 76.50 & 85.95 \\
    \midrule
    CoCoOp & 94.87 & 71.75 & 81.71 \\
    \rowcolor{LightGreen}\textbf{+ProMIM} & \textbf{94.93} & \textbf{72.01} & \textbf{81.90} \\
    \midrule
    DePT & 98.00 & 76.37 & 85.84 \\
    \rowcolor{LightGreen}\textbf{+ProMIM} & \textbf{98.20} & \textbf{76.73} & \textbf{86.15} \\
    \midrule
    TCP & 97.73 & 75.57 & 85.23 \\
    \rowcolor{LightGreen}\textbf{+ProMIM} & \textbf{98.13} & \textbf{75.77} & \textbf{85.51} \\
    \bottomrule
    \end{tabular}
    }
}
\subfloat[Food.]{
\resizebox{0.24\linewidth}{!}{
    \begin{tabular}{@{}lcc|c@{}}
    \toprule
    ViT-B/16 & Base & New & H \\
    \midrule
    CLIP & 90.10 & 91.22 & 90.66 \\
    \midrule
    CoOp & 88.33 & 82.26 & 85.19 \\
    KgCoOp & 90.50 & 91.70 & 91.09 \\
    ProGrad & 90.37 & 89.59 & 89.98 \\
    MaPLe & 90.71 & 92.05 & 91.38 \\
    PromptSRC & 90.67 & 91.53 & 91.10 \\
    \midrule
    CoCoOp & \textbf{90.70} & 91.29 & 90.99 \\
    \rowcolor{LightGreen}\textbf{+ProMIM} & 90.67 & \textbf{91.77} & \textbf{91.22} \\
    \midrule
    DePT & 90.50 & 91.60 & 91.05 \\
    \rowcolor{LightGreen}\textbf{+ProMIM} & \textbf{90.77} & \textbf{91.77} & \textbf{91.27} \\
    \midrule
    TCP & 90.57 & 91.37 & 90.97 \\
    \rowcolor{LightGreen}\textbf{+ProMIM} & \textbf{90.73} & \textbf{91.50} & \textbf{91.11} \\
    \bottomrule
    \end{tabular}
    }
}
\subfloat[FGVC.]{
\resizebox{0.24\linewidth}{!}{
    \begin{tabular}{@{}lcc|c@{}}
    \toprule
    ViT-B/16 & Base & New & H \\
    \midrule
    CLIP & 27.19 & 36.29 & 31.09 \\
    \midrule
    CoOp & 40.44 & 22.30 & 28.75 \\
    KgCoOp & 36.21 & 33.55 & 34.83 \\
    ProGrad & 40.54 & 27.57 & 32.82 \\
    MaPLe & 37.44 & 35.61 & 36.50 \\
    PromptSRC & 43.73 & 37.87 & 40.15 \\
    \midrule
    CoCoOp & 33.41 & 23.71 & 27.74 \\
    \rowcolor{LightGreen}\textbf{+ProMIM} & \textbf{35.70} & \textbf{33.63} & \textbf{34.63} \\
    \midrule
    DePT & 43.20 & 34.83 & 38.57 \\
    \rowcolor{LightGreen}\textbf{+ProMIM} & \textbf{43.56} & \textbf{34.91} & \textbf{38.76} \\
    \midrule
    TCP & 41.97 & 34.43 & 37.83 \\
    \rowcolor{LightGreen}\textbf{+ProMIM} & \textbf{42.12} & \textbf{36.83} & \textbf{39.30} \\
    \bottomrule
    \end{tabular}
    }
}
\quad%
\subfloat[SUN397.]{
\resizebox{0.24\linewidth}{!}{
    \begin{tabular}{@{}lcc|c@{}}
    \toprule
    ViT-B/16 & Base & New & H \\
    \midrule
    CLIP & 69.36 & 75.35 & 72.23 \\
    \midrule
    CoOp & 80.60 & 65.89 & 72.51 \\
    KgCoOp & 80.29 & 76.53 & 78.36 \\
    ProGrad & 81.26 & 74.17 & 77.55 \\
    MaPLe & 80.82 & 78.70 & 79.75 \\
    PromptSRC & 82.67 & 78.47 & 80.52 \\
    \midrule
    CoCoOp & 79.74 & 76.86 & 78.27 \\
    \rowcolor{LightGreen}\textbf{+ProMIM} & \textbf{79.91} & \textbf{76.93} & \textbf{78.39} \\
    \midrule
    DePT & 82.33 & 77.80 & 80.00 \\
    \rowcolor{LightGreen}\textbf{+ProMIM} & \textbf{82.43} & \textbf{77.97} & \textbf{80.14} \\
    \midrule
    TCP & 82.63 & 78.20 & 80.35 \\
    \rowcolor{LightGreen}\textbf{+ProMIM} & \textbf{82.69} & \textbf{78.35} & \textbf{80.46} \\
    \bottomrule
    \end{tabular}
    }
}
\subfloat[DTD.]{
\resizebox{0.24\linewidth}{!}{
    \begin{tabular}{@{}lcc|c@{}}
    \toprule
    ViT-B/16 & Base & New & H \\
    \midrule
    CLIP & 53.24 & 59.90 & 56.37 \\
    \midrule
    CoOp & 79.44 & 41.18 & 54.24 \\
    KgCoOp & 77.55 & 54.99 & 64.35 \\
    ProGrad & 77.35 & 52.35 & 62.45 \\
    MaPLe & 80.36 & 59.18 & 68.16 \\
    PromptSRC & 83.37 & 62.97 & 71.75 \\
    \midrule
    CoCoOp & 77.01 & 56.00 & 64.85 \\
    \rowcolor{LightGreen}\textbf{+ProMIM} & \textbf{77.47} & \textbf{57.63} & \textbf{66.09} \\
    \midrule
    DePT & 82.20 & 59.13 & 68.78 \\
    \rowcolor{LightGreen}\textbf{+ProMIM} & \textbf{83.13} & \textbf{61.60} & \textbf{70.76} \\
    \midrule
    TCP & 82.77 & 58.07 & 68.25 \\
    \rowcolor{LightGreen}\textbf{+ProMIM} & \textbf{83.63} & \textbf{60.73} & \textbf{66.94} \\
    \bottomrule
    \end{tabular}
    }
}
\subfloat[EuroSAT.]{
\resizebox{0.24\linewidth}{!}{
    \begin{tabular}{@{}lcc|c@{}}
    \toprule
    ViT-B/16 & Base & New & H \\
    \midrule
    CLIP & 56.48 & 64.05 & 60.03 \\
    \midrule
    CoOp & 92.19 & 54.74 & 68.69 \\
    KgCoOp & 85.64 & 64.34 & 73.48 \\
    ProGrad & 90.11 & 60.89 & 72.67 \\
    MaPLe & 94.07 & 73.23 & 82.35 \\
    PromptSRC & 92.90 & 73.90 & 82.32 \\
    \midrule
    CoCoOp & \textbf{87.49} & 60.04 & 71.21 \\
    \rowcolor{LightGreen}\textbf{+ProMIM} & 84.97 & \textbf{68.03} & \textbf{75.56} \\
    \midrule
    DePT & 89.03 & 71.07 & 79.04 \\
    \rowcolor{LightGreen}\textbf{+ProMIM} & \textbf{91.30} & \textbf{71.87} & \textbf{80.43} \\
    \midrule
    TCP & 91.63 & 74.73 & 82.32 \\
    \rowcolor{LightGreen}\textbf{+ProMIM} & \textbf{91.82} & \textbf{74.99} & \textbf{82.56} \\
    \bottomrule
    \end{tabular}
    }
}
\subfloat[UCF101.]{
\resizebox{0.24\linewidth}{!}{
    \begin{tabular}{@{}lcc|c@{}}
    \toprule
    ViT-B/16 & Base & New & H \\
    \midrule
    CLIP & 70.53 & 77.50 & 73.85 \\
    \midrule
    CoOp & 84.69 & 56.05 & 67.46 \\
    KgCoOp & 82.89 & 76.67 & 79.65 \\
    ProGrad & 84.33 & 74.94 & 79.35 \\
    MaPLe & 83.00 & 78.66 & 82.35 \\
    PromptSRC & 87.10 & 78.80 & 82.74 \\
    \midrule
    CoCoOp & \textbf{82.33} & 73.45 & 77.64 \\
    \rowcolor{LightGreen}\textbf{+ProMIM} & 82.23 & \textbf{76.50} & \textbf{79.26} \\
    \midrule
    DePT & 85.80 & 77.23 & 81.29 \\
    \rowcolor{LightGreen}\textbf{+ProMIM} & \textbf{85.80} & \textbf{77.87} & \textbf{81.64} \\
    \midrule
    TCP & 87.13 & 80.77 & 83.83 \\
    \rowcolor{LightGreen}\textbf{+ProMIM} & \textbf{87.83} & \textbf{80.97} & \textbf{84.26} \\
    \bottomrule
    \end{tabular}
    }
}
\end{table*}

\begin{table*}[!htp]
    \caption{Comparison in the cross-dataset transfer setting. The source model is trained on ImageNet \cite{deng2009imagenet}.}
\label{table2}
    \centering
    \setlength{\tabcolsep}{5pt}
    \begin{tabular}{@{}l|c|cccccccccc|c@{}}
    \toprule
        \multirow{2}{*}{Method} & Source & \multicolumn{11}{c}{Target} \\
        \cmidrule(r){2-2} \cmidrule{3-13}
        & ImageNet & Caltech & Pets & Cars & Flowers & Food & FGVC & SUN397 & DTD & EuroSAT & UCF101 & \textbf{Avg.} \\
        \midrule
        CoOp & 71.51 & 93.70 & 89.14 & 64.51 & 68.71 & 85.30 & 18.47 & 64.15 & 41.92 & 46.39 & 66.55 & 63.88 \\
        \midrule
        CoCoOp & \textbf{71.02} & \textbf{94.43} & 90.14 & \textbf{65.32} & \textbf{71.88} & 86.06 & 22.94 & 67.36 & 45.73 & 45.37 & \textbf{68.21} & 65.74 \\
        \rowcolor{LightGreen}\textbf{+ProMIM} & 70.60 & 93.97 & \textbf{90.33} & 65.03 & 71.47 & \textbf{86.67} & \textbf{23.87} & \textbf{67.40} & \textbf{47.60} & \textbf{46.77} & 67.80 & \textbf{66.09} \\
        \midrule
        DePT & \textbf{72.77} & 94.10 & \textbf{90.63} & \textbf{66.23} & \textbf{72.17} & 86.27 & 22.90 & \textbf{67.30} & 45.50 & 44.17 & \textbf{69.53} & 65.88\\
        \rowcolor{LightGreen}\textbf{+ProMIM} & 72.53 & \textbf{94.13} & 90.03 & 65.40 & 69.97 & \textbf{86.50} & \textbf{23.70} & 66.83 & \textbf{46.80} & \textbf{49.85} & 68.13 & \textbf{66.14} \\
        \midrule
        TCP & \textbf{71.40} & 93.97 & 91.25 & 64.69 & \textbf{71.21} & 86.69 & 23.45 & \textbf{67.15} & 44.35 & 51.45 & 68.73 & 66.29\\
        \rowcolor{LightGreen}\textbf{+ProMIM} & 71.23 & \textbf{94.03} & \textbf{91.33} & \textbf{65.20} & 71.11 & \textbf{86.86} & \textbf{23.79} & 66.64 & \textbf{46.20} & \textbf{51.85} & \textbf{68.93} & \textbf{66.59} \\
    \bottomrule
    \end{tabular}
\end{table*}

\paragraph{\textbf{Datasets}}
For the initial two configurations, namely base-to-new generalization and cross-dataset transferability, we utilized the 11 image recognition datasets detailed in \cite{zhou2022learning}, which span a broad spectrum of recognition tasks. This benchmark includes ImageNet \cite{deng2009imagenet} and Caltech101 (Caltech) \cite{fei2004learning} for generic object classification; OxfordPets (Pets) \cite{parkhi2012cats}, StanfordCars (Cars) \cite{krause20133d}, Flowers102 (Flowers) \cite{nilsback2008automated}, Food101 (Food) \cite{bossard2014food}, and FGVCAircraft (FGVC) \cite{maji2013fine} for fine-grained classification; SUN397 \cite{xiao2010sun} for scene recognition; UCF101 \cite{soomro2012ucf101} for action recognition; DTD \cite{cimpoi2014describing} for texture classification; and EuroSAT \cite{helber2019eurosat} for satellite image recognition.
In the domain generalization setup, we used ImageNet as the source dataset, with four domain-shifted ImageNet variants serving as target datasets: ImageNet-V2 \cite{recht2019imagenet}, ImageNet-Sketch \cite{wang2019learning}, ImageNet-A \cite{hendrycks2021natural}, and ImageNet-R \cite{hendrycks2021many}. Following the protocol described in \cite{zhou2022learning}, we randomly selected a few-shot training subset from each dataset, utilizing the original test set for evaluation. To validate the effectiveness of our approach, we conducted experiments with the maximum shot count considered in \cite{zhou2022learning}, specifically 16-shot, and reported results averaged over three independent runs.

\paragraph{\textbf{Baselines}}
We compare our ProMIM against several state-of-the-art methods, including zero-shot and linear probe CLIP \cite{radford2021learning}, CoOp \cite{zhou2022learning}, CoCoOp \cite{zhou2022conditional}, DePT \cite{zhang2024dept}, and TCP \cite{yao2024tcp}.
For a fair comparison, the baseline results are directly sourced from their respective original publications.

\paragraph{\textbf{Implementation details}}
Our implementation builds upon the CoCoOp codebase, utilizing CLIP (ViT-B/16) \cite{radford2021learning} as the pre-trained vision-language model for evaluation. We fine-tune the model in a few-shot setting with 16 samples per class, employing an SGD optimizer over 10 epochs with a batch size of 1 and a learning rate of 0.02. To ensure consistency, we limit the number of context tokens to 4. We report the average accuracy over three independent runs for each experimental setup. The training and evaluation processes are performed on a single NVIDIA H100 GPU, completing within a day for all 11 datasets. The hyperparameter $\lambda$ is set to 2.0.

\subsection{Generalization From Base to New Classes}
The main focus of this study is to solve the problem of overfitting in CoCoOp methods.
Following the previous methods \cite{zhou2022learning,zhou2022conditional,yao2023visual,zhu2023prompt,khattak2023maple,zhang2024dept,yao2024tcp}, we split each data set into the \textit{Base} and \textit{New} classes.
The models are trained on the \textit{Base} classes and evaluated on the test sets of the \textit{Base} and \textit{New} classes.
The detailed results are presented in Table~\ref{table1}.

As demonstrated in Table~\ref{table1}, ProMIM achieves a 73.96\% accuracy, significantly outperforming CoOp and CoCoOp on the \textit{New} benchmarks.
This improvement can be attributed to ProMIM’s ability to mitigate data leakage by utilizing masked image features to generate context tokens, which in turn enhances generalization to unseen classes.
Although ProMIM’s performance on the \textit{Base} setting is slightly lower, due to the additional regularization applied to reduce overfitting and improve generalization, it demonstrates substantial gains over previous state-of-the-art methods across challenging datasets, including ImageNet, StanfordCars, Oxford Flowers, FGVCAircraft, DTD, and EuroSAT, where there remains optimization potential.
Notably, applying ProMIM to enhance DePT and TCP results in further performance improvements, with accuracies of 84.12\% for \textit{Base} classes and 75.54\% for \textit{New} classes with DePT and 84.41\% and 76.05\% for \textit{Base} and \textit{New} classes with TCP, respectively.
This enhanced performance underscores the plug-and-play versatility of the proposed ProMIM method, facilitating seamless integration with existing methods.

\begin{table}[!htp]
    \caption{Comparison of manual and learning-based prompts in domain generalization.}
\label{table3}
    \centering
    \resizebox{\linewidth}{!}{
    \begin{tabular}{l|c|cccc|c}
    \toprule
         \multirow{2}{*}{Method} & Source & \multicolumn{5}{c}{Target} \\
         \cmidrule(r){2-2} \cmidrule{3-7}
         & ImageNet & -V2 & -Sketch & -A & -R & \textbf{Avg.} \\
         \midrule
        CLIP & 66.73 & 60.83 & 46.15 & 47.77 & 73.96 & 57.17 \\
        \midrule
        CoOp & 71.51 & 64.20 & 47.99 & 49.71 & 75.21 & 59.28 \\
        \midrule
        CoCoOp & \textbf{71.02} & 64.07 & 48.75 & 50.63 & 76.18 & 59.90 \\
        \rowcolor{LightGreen}\textbf{+ProMIM} & 70.60 & \textbf{64.12} & \textbf{48.84} & \textbf{50.85} & \textbf{76.63} & \textbf{60.11} \\
        \midrule
        DePT & - & - & - & - & - & -\\
        \rowcolor{LightGreen}\textbf{+ProMIM} & \textbf{72.53} & 63.97 & 45.67 & 46.30 & 73.00 & 57.23 \\
        \midrule
        TCP & 71.20 & 64.60 & 49.50 & \textbf{51.20} & 76.73 & 60.51 \\
        \rowcolor{LightGreen}\textbf{+ProMIM} & \textbf{71.23} & \textbf{64.65} & \textbf{49.51} & 51.11 & \textbf{77.01} & \textbf{60.57} \\
    \bottomrule
    \end{tabular}
    }
\end{table}

\begin{table*}[h]
    \caption{Comparison of average performance across all 11 datasets in the base-to-new setting with varying $K$-shot samples.}
\label{table4}
    \centering
    \setlength{\tabcolsep}{8pt}
    \begin{tabular}{l|l|ccc||ccc||ccc}
    \toprule
        \multirow{2}{*}{Backbones} & \multirow{2}{*}{Methods} & \multicolumn{3}{c||}{K=4} & \multicolumn{3}{c||}{K=8} & \multicolumn{3}{c}{K=16} \\
        \cmidrule{3-11}
        & & Base & New & H & Base & New & H & Base & New & H \\
        \midrule
        \multirow{5}{*}{ViT-B/16} & CoOp & 78.43 & 68.03 & 72.44 & \textbf{80.73} & 68.39 & 73.50 & \textbf{82.63} & 67.99 & 74.60 \\
        & CoCoOp & 76.72 & 73.34 & 74.85 & 78.56 & 72.00 & 74.90 & 80.47 & 71.69 & 75.83 \\
        & ProGrad & 79.18 & 71.14 & 74.62 & 80.62 & 71.02 & 75.20 & 82.48 & 70.75 & 76.16 \\
        & KgCoOp & \textbf{79.92} & 73.11 & \textbf{75.90} & 78.36 & 73.89 & 76.06 & 80.73 & 73.60 & 77.00 \\
        \cmidrule{2-11}
        \rowcolor{LightGreen} & \textbf{ProMIM} & 77.87 & \textbf{76.10} & 75.62 & 78.20 & \textbf{74.50} & \textbf{76.30} & 80.64 & \textbf{73.96} & \textbf{77.16} \\
    \bottomrule
    \end{tabular}
\end{table*}

\subsection{Cross-Dataset Transferability Evaluation}
To further assess the generalization capability of the ProMIM framework, we conduct a cross-dataset evaluation by training all models on ImageNet and directly testing on 10 additional downstream datasets. The comparative results between ProMIM and existing methods are presented in Table~\ref{table2}. 
As shown in Table~\ref{table2}, ProMIM achieves higher average performance than previous prompt-based methods, including CoOp \cite{zhou2022learning}, CoCoOp \cite{zhou2022conditional}, and DePT \cite{zhang2024dept}, which do not integrate prompts deep within the architecture, as seen in methods like MaPLe \cite{khattak2023maple} or TCP \cite{yao2024tcp}.
This demonstrates the effectiveness of ProMIM in learning general knowledge.
For instance, despite having only 70.60\% accuracy on ImageNet, ProMIM achieves an average accuracy of 66.09\% across 10 datasets, higher than CoOp, CoCoOp, DePT, and TCP.
Furthermore, when DePT and TCP are used with ProMIM, we achieve an average performance higher than those of the existing methods with accuracy improvements of 0.26\% and 0.30\% overall, respectively.

\subsection{Domain Generalization}
The domain generalization results are summarized in Table~\ref{table3}. Here, the original ImageNet dataset is used as the source for model fine-tuning, followed by the evaluation of four distinct ImageNet variants, each representing a different distribution. As seen in Table~\ref{table3}, ProMIM exhibits slightly lower performance on ImageNet but effectively aligns with performance levels on the target datasets. Notably, when combined with DePT \cite{zhang2024dept}, ProMIM achieves peak performance on ImageNet; however, this pairing results in a substantial drop in generalization, indicating a reduced capacity of DePT to generalize across a broader range of unseen classes. In contrast, ProMIM combined with TCP attains the highest overall performance over 4 target datasets.

\section{Ablation study}
\subsection{Few-shot Classification}
To evaluate the effectiveness of the proposed ProMIM, we benchmark its performance against previous prompt-based methods, including CoOp \cite{zhou2022learning}, CoCoOp \cite{zhou2022conditional}, KgCoOp \cite{yao2023visual}, and ProGrad \cite{zhu2023prompt}, across various backbone models and different $K$-shot sample settings. Specifically, we utilize  ViT-B/16, a transformer-based deep learning backbone, as the visual encoder for image feature extraction. Additionally, we conduct experiments under three few-shot configurations: 4-shot, 8-shot, and 16-shot. The averaged results are summarized in Table~\ref{table4}.
Generally, ProMIM surpasses all methods for the performance on the \textit{New} classes. In particular, ProMIM achieves best performance on \textit{New} classes despite the number of shots.
Especially when the number of shots increases, ProMIM surpasses all other methods.

\begin{table}[!ht]
    \caption{Comparison of average performance across all 11 datasets in the base-to-new setting, with and without the inclusion of MIM-context and $\mathcal{L}_{kg}$.}
\label{table5}
    \centering
    \setlength{\tabcolsep}{7pt}
    \begin{tabular}{l|c|c|ccc}
    \toprule
        \multirow{2}{*}{Methods} & MIM & \multirow{2}{*}{$\mathcal{L}_{kg}$} & \multicolumn{3}{c}{Avg 11 datasets} \\
        \cmidrule{4-6}
        & context & & Base & New & H \\
        \midrule
        CoOp & \ding{55} & \ding{55} & \textbf{82.63} & 67.99 & 74.60 \\
        CoCoOp & \ding{55} & \ding{55} & 80.47 & 71.69 & 75.83 \\
        ProGrad & \ding{55} & \ding{55} & 82.48 & 70.75 & 76.16 \\
        KgCoOp & \ding{55} & \ding{51} & 80.73 & 73.60 & 77.00 \\
        \midrule
        \rowcolor{LightGreen} \textbf{ProMIM} & \ding{51} & \ding{55}  & 80.40 & 72.32 & 76.15 \\
        \rowcolor{LightGreen} \textbf{ProMIM} & \ding{51} & \ding{51} & 80.64 & \textbf{73.96} & \textbf{77.16} \\
    \bottomrule
    \end{tabular}
\end{table}

\subsection{Impact of each module}
Our work focuses on utilizing the features from masked images as a condition for prompt learning, referred to as the MIM context.
This approach aims to address the overfitting issue caused by data leakage in CoCoOp method \cite{zhou2022conditional}. 
As demonstrated in Table~\ref{table5}, the inclusion of the MIM context significantly enhances performance compared to CoOp, CoCoOp, KgCoOp, and ProGrad, particularly in the \textit{New} and \textit{H} metrics.
Notably, for the \textit{New} performance, incorporating the MIM context results in an improvement of almost 2\% over ProGrad.
Furthermore, when we apply $\mathcal{L}_{kg}$ from KgCoOp, the \textit{New} performance exceeds that of KgCoOp, achieving 73.96\% compared to 73.60\%.
This superior performance further validates the effectiveness of using masked image features as a condition, as it helps to prevent overfitting and enhances overall performance over 11 datasets.



\subsection{Computational Cost}
The computational cost of our ProMIM is shown in Table~\ref{cost}. 
As observed, the additional computational cost is low (even \textit{negligible}) compared to the performance improvement established by ProMIM. Especially since no additional modules are introduced, the learnable parameters are unchanged.

\begin{table*}[t]
\centering
\caption{Ablation study on computational cost and masking strategy.}
    \subfloat[
    Computational cost of ProMIM. “ms”: millisecond per image. Experiments are performed on an H100 GPU.
    \label{cost}
    ]{
    \centering
    \begin{minipage}{0.7\linewidth}{
    \begin{center}
        \begin{tabular}{l|cccc|c}
        \toprule
          Method & Learnable Parameters & Training time & Inference time & Memory & H (\textit{avg}) \\
            \midrule
            CoOp& 8K &105min &2.78ms &11264MB  &71.66\\
            \midrule            
            KgCoOp& 2K &106min &2.79ms &11254MB &77.00\\
            \midrule            
            MaPLe& 3635K &111min &2.75ms &11048MB &78.55\\
            \midrule            
            CoCoOp&69K &420min &127.06ms &11310MB & 75.83\\
           \textbf{+ProMIM}&69K   &\color{red}+3min\color{black} &\color{red}+0.1ms\color{black} &\color{red}+2MB\color{black} &\textbf{77.16} (\color{red}{+0.35}\color{black})\\
            \bottomrule
        \end{tabular}
    \end{center}}
    \end{minipage}
    }
\quad
    \subfloat[
    We compare different masking strategies.
    \label{tab:mask_types}
    ]{
    \begin{minipage}{0.25\linewidth}{
    \begin{center}
    \centering
    \begin{tabular}{lcc}
    \toprule
    case & ratio & H \textit{(avg)} \\
    \midrule
    random & 0.50 & 76.42 \\
    random & 0.75 & \textbf{77.16} \\
    block & 0.50 & 76.31 \\
    block & 0.75 & 75.12 \\
    \bottomrule
    \end{tabular}
    \end{center}}
    \end{minipage}
    }
\end{table*}

\subsection{\texorpdfstring{Mask ratio and $\lambda$ value}{Mask ratio and lambda value}}
We thus confirm the rationale and effectiveness of ProMIM’s design, with the relevant results presented in Table~\ref{table7} and Table~\ref{table8}. 
Notably, unlike KgCoOp, ProMIM achieves optimal performance at $\lambda = 2.0$, indicating that masked image features provide robust regularization. 
This suggests the need to minimize $\lambda$ for ProMIM, whereas KgCoOp requires a larger $\lambda$ to effectively regularize the loss. 
Furthermore, when a significant portion of the input image is conditioned, performance declines, likely due to data leakage; the best results are observed when 95\% of the input image is masked.

\begin{table}[!h]
    \caption{The quantitative analysis of mask ratio on average of 11 datasets.}
\label{table7}
    \centering
    \setlength{\tabcolsep}{8pt}
    \begin{tabular}{@{}l|ccccc@{}}
    \toprule
        mask ratio (\%) & 0.25 & 0.5 & \textbf{0.75} & 0.95 & 0.99 \\
        \midrule
        H & 76.21 & 76.42 & \textbf{77.16} & 75.36 & 74.72 \\
    \bottomrule
    \end{tabular}
\end{table}

\begin{table}[!h]
    \caption{Quantitative analysis of $\mathcal{L}_{kg}$ for varying $\lambda$ values, averaged over 11 datasets.}
    \label{table8}
    \centering
    \setlength{\tabcolsep}{6pt}
    \begin{tabular}{l|ccccccc}
    \toprule
        $\lambda$ & 0.0 & 1.0 & \textbf{2.0} & 4.0 & 6.0 & 8.0 & 10.0 \\
        \midrule
        H & 76.15 & 76.58 & \textbf{77.16} & 76.87 & 76.37 & 76.53 & 76.57 \\
    \bottomrule
    \end{tabular}
\end{table}

\subsection{Mask sampling strategy.}
In Table~\ref{tab:mask_types}, we analyze mask sampling strategies. The \textit{block-wise} strategy, introduced by \cite{bao2021beit}, typically eliminates large contiguous sections. Our ProMIM, when using block-wise masking, performs adequately at a 50\% ratio, but its effectiveness diminishes at 75\%. This approach proves more difficult than random sampling due to an observed increase in training loss. Simple random sampling yields the best results for our ProMIM, as it supports a higher masking ratio, offering substantial speedup and maintaining high accuracy.

\begin{figure}[!ht]
  \centering
  \begin{subfigure}{\linewidth}
    \includegraphics[width=\linewidth]{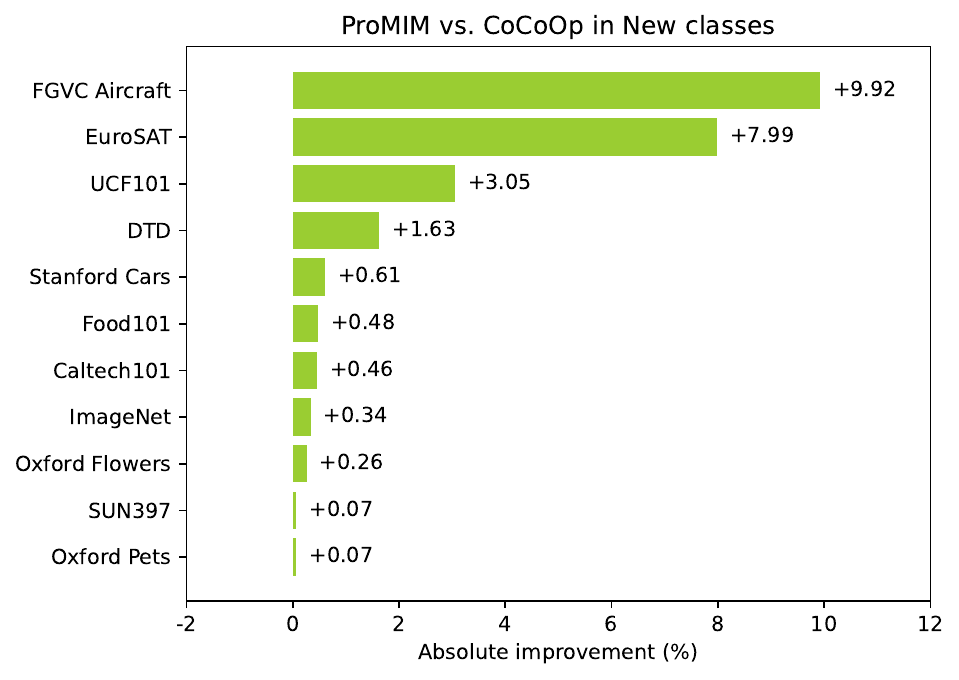}
    \caption{ProMIM consistently outperforms CoCoOp on unseen classes across all datasets.}
    \label{fig:new-gain}
  \end{subfigure}
  \begin{subfigure}{\linewidth}
    \includegraphics[width=\linewidth]{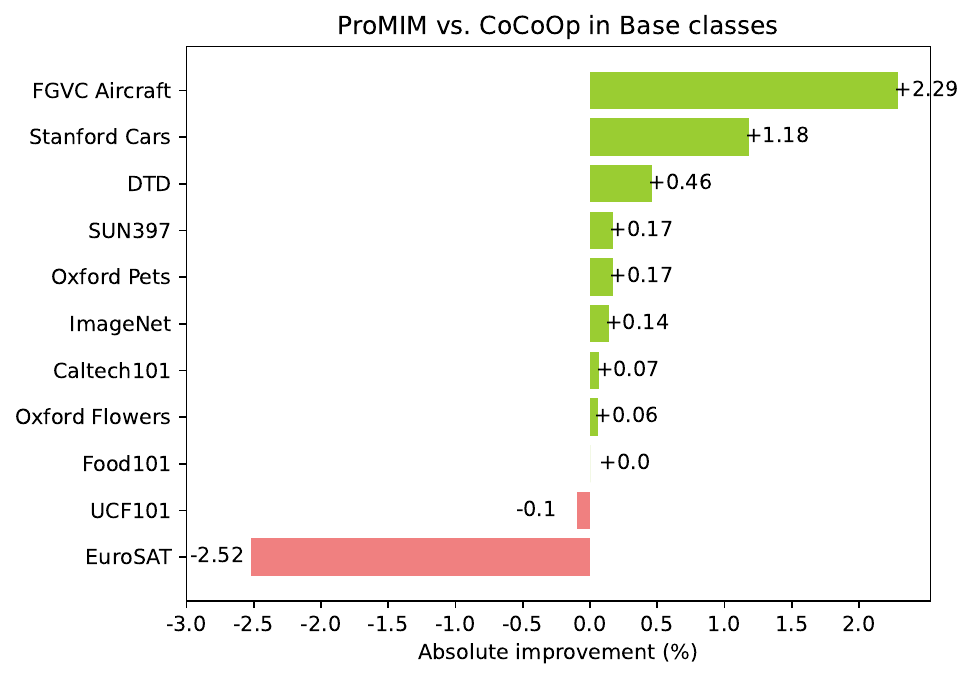}
    \caption{ProMIM reduces the overfitting effect, resulting in lower base performances.}
    \label{fig:base-gain}
  \end{subfigure}
  \caption{Extensive evaluation of ProMIM versus CoCoOp in the context of base-to-new class generalization.}
  \label{fig:gain}
\end{figure}

\subsection{ProMIM enhances the generalization toward unseen classes}
Figure~\ref{fig:gain} illustrates the absolute performance gains of ProMIM over CoCoOp, our primary comparison baseline. The addition of regularization constraints in ProMIM results in a slight reduction in accuracy on \textit{Base} classes, as expected, due to the model’s reduced reliance on memorized class-specific features. 
However, we observe a consistent improvement in accuracy on 10 out of 11 datasets, especially with more than 10\% improvement on the FGVCAircraft dataset.
These results underscore the ProMIM's capability to debias the model from overfitting on \textit{Base} classes, thereby enhancing its adaptability and generalization capability on unseen categories.

\section{Conclusion}
Common CoOp-based prompt learning methods tend to overfit to classes encountered during fine-tuning, limiting their generalization to novel, unseen classes. 
 To address this challenge, we introduce Masked Image Modeling-guided Conditional Prompt Learning (ProMIM), a flexible, plug-and-play framework that enhances prompt learning by integrating masked image modeling (MIM) into existing vision-language model (VLM) pipelines. ProMIM utilizes MIM’s robust masking strategy to generate instance-conditioned prompts, effectively improving feature robustness and mitigating overfitting. This approach prevents data leakage from the visual to the textual branch, ensuring that the model establishes a stronger contextual framework for generating prompts.
Evidence from various benchmarks and tasks confirms ProMIM's effectiveness and versatility, demonstrating its potential as a lightweight, practical solution for real-world vision-language applications.

\begin{acks}
This work was partly supported by the Korea government (MSIT), IITP, Korea, under the ICT Creative Consilience program (IITP-2025-RS-2020-II201821,30\%), Development of Brain Disease (Stroke) Prediction Model based on Fundus Image Analysis (RS-2024-00459512, 30\%), AI Innovation Hub (RS-2021-II212068, 20\%), and AI Graduate School Program (Sungkyunkwan University, RS-2019-II190421, 20\%).
\end{acks}

\bibliographystyle{ACM-Reference-Format}

\end{document}